\documentclass{article}

\usepackage{PRIMEarxiv}

\usepackage{hyperref}
\hypersetup{
  colorlinks   = true, 
  urlcolor     = blue, 
  linkcolor    = blue, 
  citecolor   = blue 
}

\usepackage[utf8]{inputenc} 
\usepackage[T1]{fontenc}    
\usepackage{hyperref}       
\usepackage{url}            

\usepackage{booktabs}       
\usepackage{amsfonts}       
\usepackage{nicefrac}       
\usepackage{microtype}      
\usepackage{lipsum}
\usepackage{fancyhdr}       
\usepackage{graphicx}       
\graphicspath{{media/}}     
\usepackage[numbers]{natbib}
\usepackage{doi}

\title{Taxonomy to Regulation: A (Geo)Political Taxonomy for AI Risks and Regulatory Measures in the EU AI Act
}

\author{
Sinan Arda,\\
  University of North Carolina at Chapel Hill \\
  Chapel Hill, NC, USA\\
  \texttt{sinanarda@alumni.unc.edu} \\
}

\begin{document}
\maketitle
\begin{abstract}
Technological innovations have shown remarkable capabilities to benefit and harm society alike. AI constitutes a democratized sophisticated technology accessible to large parts of society, including malicious actors. This work proposes a taxonomy focusing on on (geo)political risks associated with AI. It identifies 12 risks in total divided into four categories: (1) Geopolitical Pressures, (2) Malicious Usage, (3) Environmental, Social, and Ethical Risks, and (4) Privacy and Trust Violations. Incorporating a regulatory side, this paper conducts a policy assessment of the EU AI Act. Adopted in March 2023, the landmark regulation has the potential to have a positive top-down impact concerning AI risk reduction but needs regulatory adjustments to mitigate risks more comprehensively. Regulatory exceptions for open-source models, excessively high parameters for the classification of GPAI models as a systemic risk, and the exclusion of systems designed exclusively for military purposes from the regulation’s obligations leave room for future action.

\end{abstract}


\section{Introduction}
AI has demonstrated great potential to benefit society in numerous areas, including medicine and scientific research. Simultaneously, the capabilities of present-day AI, alongside forecasts on future AI progress, have rightfully stimulated calls for increased government oversight across the policy spectrum to ensure responsible AI development and usage \cite{anderljung2023frontier}. After the launch of ChatGPT in November 2022 by the Californian tech firm OpenAI, the technology reached a fever pitch \cite{theeconomistAIModelsWill2023}. Contemporary models have demonstrated an unprecedented ability to understand and generate human-like output. Linked to the rapid proliferation of AI and the technology's arrival in the public in 2022, concerns over the risks of AI have intensified. Meanwhile, the question of AI governance has become a policy priority and led to political agreements. In October 2023, President Biden issued an Executive Order intending to manage AI risks \cite{thewhitehouseFACTSHEETPresident2023}. Two months later, the EU, after over two years of drafting time and inter-institutional negotiations, has reached an initial political agreement on the EU Artificial Intelligence Act (AI Act), which is said to serve as the world's first comprehensive regulatory example for AI governance \cite{clarahainsdorfDawnEUAI2023}. This work captures the risk-regulation dynamic in AI research by focusing specifically on the political risks of AI and regulatory action taken under the AI Act.

While AI creates benefits, it is a double-edged sword as it also poses significant harms that have to be addressed. This work is motivated by the ongoing widespread use of AI models, powered by constant technological progress, and a strong belief in an interdisciplinary effort to mitigate contemporary and future risks for our societies. As the objective of this work is two-fold, conceptualizing political AI risks and conducting a regulatory analysis of the EU response to the challenges linked to AI proliferation, two research questions guide the research design: (RQ1) What kind of risks does AI pose for domestic and international politics? (RQ2) How does the European Union Artificial Intelligence Act mitigate the risks associated with AI?

The dual objective of conceptualizing political AI risks and assessing the AI Act will be based on a qualitative risk and policy analysis of empirical material. To answer RQ1, a (geo)political risk taxonomy will be established that captures AI's implications for domestic and international politics. The proposed taxonomy incorporates contemporary findings in AI risk research and narrows down the complex risk environment to the political domain. In the case of RQ2, the EU's legislative process toward the AI Act will be analyzed by referring to the first three stages of the policy cycle. Afterward, the regulatory loopholes of the present agreement will be discussed. Utilized qualitative materials include scholarly work on EU policy-analysis frameworks, interdisciplinary findings on the risk linked to AI, official legislative material published by the EU institutions on the AI Act, and commentaries on the process of institutional negotiations. 

The structure of this work is outlined as follows: Chapter 2 will explore policy-making in the EU, focusing on the policy cycle as the conceptual framework. It is selected as the relevant concept of analyzing policy responses because this work is interested in the AI Act as a legislative policy response to the potential disruptive risks of AI. Due to the objective of this research project, answering the constructed research questions follows a two-stage approach (Chapters 2 and 3). Chapter 3 will answer RQ1. It will commence with a literature review to map the present state of the art of existing AI risk taxonomies and to systematically identify risks. Subsequently, a (geo)political taxonomy of AI risk will be proposed. The "Danger of an AI Arms Race and Artificial Weaponry Systems" as well as "Deepfakes and the Proliferation of Disinformation" will be assessed in detail. Chapter 4 will answer RQ2. Therefore, the AI Act will be introduced and analyzed, encompassing the Proposal put forth by the European Commission in 2021 and the preliminary agreement reached in December 2023. The legislative analysis supports a subsequent assessment of the legislation's regulatory loopholes and extraterritorial spill-over potential. Chapter 5 will summarize this work's main findings and assess its implications for future work concerning AI governance.

\section{EU Public Policy Analysis}
Regulatory policy-making in the EU occurs in the context of the varied regulatory interests and traditions of Member States. Because of the heterogeneity of domestic regulations and policy preferences, the bloc's regulatory policy-making has been described as a "policy patchwork" \cite{heritierAccommodationDiversityEuropean1996} in which different regulatory approaches are linked to a single policy outcome. Due to the absence of a coherent regulatory framework in such policy patchwork, Member States are likely to face costly policies that they have to "download" from the European to the national level \cite{borzelPaceSettingFootDraggingFenceSitting2002}. Europeanization is a two-way process that entails both a "bottom-up" and "top-down" dimension. While the first dimension describes the evolution of the European institutions as a set of new norms, rules, and practices, the second one describes the institution's impact on the political structures and processes of the Member States. A policy fit between the EU and national level reduces implementation costs. Consequently, Member States hold an incentive to upload their domestic policies to the EU level to minimize future downloading costs. 

Börzel conceptualized Member State responses to Europeanization as three distinguishable strategies: pace-setting, i.e., pushing policies at the European level that display a Member State's domestic policies and reduce implementation costs; foot-dragging, i.e., blocking or delaying costly policies to either prevent them or receive compensation for linked implementation costs; and fence-setting, i.e., neither pushing policies nor blocking them but building coalitions with both pace-setters and foot-draggers \cite{borzelPaceSettingFootDraggingFenceSitting2002}. As the AI Act is a landmark Europeanization of AI, it can be expected - based on ex-ante cost assessments \cite{laurerClarifyingCostsEU2021} - that the legislation will result in expensive implementation costs at the national level. Consequently, the policy-making process of the AI Act has the potential to resemble the conceptualized Member State responses discussed above (pace-setting, foot-dragging, and fence-setting). 

Conducting conceptualized research on public policy, such as the AI Act, can be pursued through a range of approaches. The so-called policy-cycle represents the most common public policy research framework \cite{olsen}. Its well-established heuristic can be utilized to conceptually understand and break down the EU policy-making process. Yet, EU polity differs from the national level, leading to significant implications for its policy-cycle \cite{heidbrederEUPolicyProcess2018}. The stages of the policy-cycle are: problem definition and agenda-setting, policy formulation, decision making, policy implementation, and policy evaluation (and termination). Although the number of stages has been widely debated without reaching a definite resolution, the listed five stages are typically evident in policy research and have become the conventional approach \cite{heidbrederEUPolicyProcess2018, wegrichTheoriesPolicyCycle2006,olsen}. As the EU AI Act is yet to be implemented, only the first three stages are of interest, i.e., problem definition and agenda setting, policy formulation, and decision-making. Authority for the first stage rests with the European Commission and the European Council, which constitute the formal agenda-setters. Once a problem has been identified, an adequate policy as a response must be formulated in the cycle's second stage. Thereby, the European Commission has the exclusive power to propose new legislation in the EU \cite{olsen, heidbrederEUPolicyProcess2018}. While green and white papers of the European Commission are considered part of the agenda-setting process, a proposal is part of the policy formulation. However, the Commission does not hold responsibility alone in formulating legislative proposals, given the input it receives from national governments, public opinion, lobbying groups, and think tanks. Decision-making traditionally occurs under the ordinary legislative procedure, established by Article 294 of the Treaty on the Functioning of the European Union. It  grants the Council of the EU and the European Parliament equal footing in co-legislating the Commission's proposal
\cite{heidbrederEUPolicyProcess2018}. The ordinary legislative procedure begins with a legislative proposal by the Commission and consists of up to three readings by the Council of the EU and the European Parliament. The Council of the EU and the European Parliament adopt a proposal either at the first  or at the second reading. However, if both co-legislators cannot reach an agreement after the second reading, a conciliation committee is convened. Conciliation is the final stage of the ordinary legislative procedure and consists of negotiations between both parties with the objective of reaching a joint text, constituting an inter-institutional agreement that has to be confirmed by both the Parliament and the Council \cite{counciloftheeuropeanunionOrdinaryLegislativeProcedure2024, europeanparliamentOrdinaryLegislativeProcedure2024}. At any stage of the procedure, so-called trilogues may be organized which are informal tripartite inter-institutional meetings between the representatives of the Parliament, the Council, and the Commission. The objective of trilogues is to reach a provisional agreement that is acceptable to both the Parliament and the Council as the co-legislators \cite{europeanparliamentInterinstitutionalNegotiations2024}.

While the EU's regulatory policies impact Member States as they are obliged to download legislation from the European to the national level, the bloc's regulatory arm can reach further. Columbia scholar Anu Bradford coined the \textit{Brussels Effect} and argued that EU legislation influences the adoption of similar laws in third countries \cite{Bradford2019}. Accordingly, the EU is able to regulate global markets by setting standards that shape the international business environment. Market forces alone are often sufficient to lead to a Europeanization of global commerce as international corporations voluntarily comply with EU law when governing their global operations. The EU's General Data Protection Regulation (GDPR) has become a well-known empirical example of the Brussels Effect as it emerged as the default for data privacy regulation, while the AI Act's potential for a similar effect has been playing an important role in the legislative debates surrounding AI \cite{siegmannBrusselsEffectArtificial2022a, englerEUAIAct2022}.




\section{A Geo-Political AI Risk Taxonomy}
\label{Chapter 3}

AI creates exciting opportunities. Simultaneously, it can (and already has in multiple instances) lead to significant harm. AI risks have attracted significant attraction, especially within academia and policymaker circles. This section assesses how the technology poses significant risks from a political perspective, contributing to an increasingly interdisciplinary research field. The assessment of AI risks is conducted in a three-step process. Firstly, the risk landscape of AI will be mapped through a literature review of existing risk taxonomies. Secondly, a risk taxonomy for risks that can be situated in the political setting is proposed. Thirdly, selected risks of the proposed taxonomy will be analyzed in detail.

\subsection{Literature Review}
\vspace{-0.15cm}
Mapping AI risks poses numerous challenges due to the wide range of applicability of existing and future AI systems. Categorizing risks can provide a way to mitigate such problems and can narrow down the risk landscape for specific research purposes. Risks have been mapped out in different taxonomies, and the scholarship greatly increased as AI arrived in the public discourse. Taxonomies, often used interchangeably with the term typologies, are risk categorizations that are conceptually or empirically derived. Primarily grounded in empirical data, a taxonomy structures previously identified risks and supports identifying new ones. Developing a risk taxonomy involves initially the collection of identified risks and sources, followed by a structuring and screening process to identify potential gaps \cite{koesslerRiskAssessmentAGI2023}. The concrete development of a risk taxonomy tends to be very context-specific and varies as researchers and organizations explore different areas of interest. This review encompasses academic papers or online reports meeting at least one of three criteria: (i) the establishment of a risk taxonomy of AI, (ii) a review of past taxonomies of AI risks, or (iii) specialist knowledge on a specific form of AI risk without necessarily taxonomizing it. Both anticipated and observed risks are included. The reader is encouraged, if of interest, to select individual pieces of this literature review to go in-depth into a specific AI risk domain as it contains research that does not solely focus on the political effects. The already-established taxonomies to be introduced in the following focus, among others, on generative AI  such as language models, large-scale societal risks, and risks in the international security setting. 

Researchers affiliated with Google's AI firm DeepMind published multiple papers dealing with AI risks. A taxonomy developed in 2022 focused on ethical and social risks associated with language models (LMs), particularly large-language models (LLMs). Twenty-one risks have been identified through a mixed methods approach that included horizon-scanning workshops and an interdisciplinary literature review incorporating computer science, linguistics, and social science. The taxonomy categorizes risks into six thematic domains: 1. Discrimination, Hate speech, and Exclusion, 2. Information Hazards, 3. Misinformation Harms, 4. Malicious Uses, 5. Human-Computer Interaction Harms, and 6. Environmental and Socioeconomic harms \cite{weidingerTaxonomyRisksPosed2022}. In 2023, DeepMind researchers proposed a three-layered evaluation framework to ensure the safety of generative AI systems (Table \ref{table1}). To provide a basis for mapping the existing safety evaluation landscape, a synthesized taxonomy of harm was aggregated based on existing literature on societal, ethical, and other risks of generative AI. Their holistic risk taxonomy of generative AI systems incorporates findings of the 2022 LLMs assessment and focuses on: 1. Representation \& Toxicity Harms, 2. Misinformation Harms, 3. Information \& Safety Harms, 4. Malicious Use, 5. Human Autonomy \& Integrity Harms, 6. Socioeconomic \& Environmental Harms \cite{weidingerSociotechnicalSafetyEvaluation2023}. Scholars at the University of Edinburgh \cite{birdTypologyRisksGenerative2023} extended and narrowed down the research conducted on generative AI models by focusing on the risks of text-to-image (TTI) generative AI models. Their non-mutually exclusive typology divides immediate risks associated with TTI-induced AI models into three key categories: 1. Discrimination and Exclusion, 2. Misuse, 3. Misinformation and Disinformation. Furthermore, it identifies involved stakeholders (developers, users, regulators, and affected parties), the nature of the potential harm (e.g., representational or allocative), and offers a distinction between anticipated and observed risks.

\begin{figure} [h]
    \centering
    \includegraphics[width=1\linewidth]{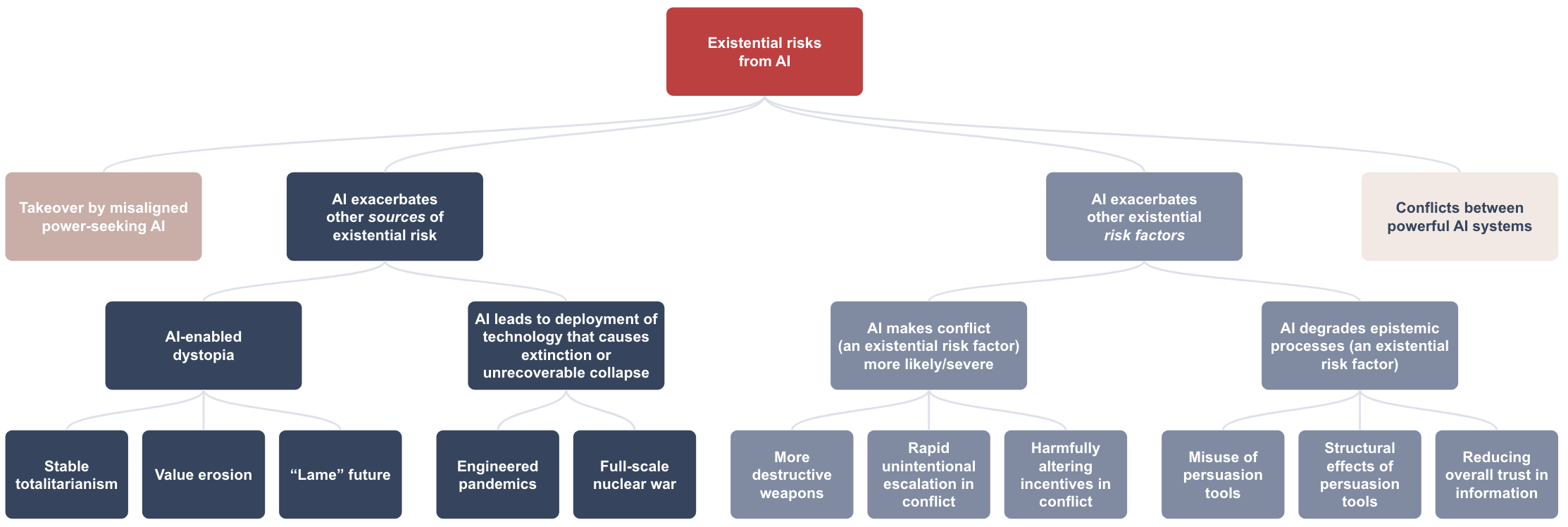}
    \caption{Typology of existential risks from AI \cite{clarkeClassifyingSourcesAI2022}}
    \label{fig3}
\end{figure}

Clarke \cite{clarkeClassifyingSourcesAI2022} developed a classification of potential sources of AI x-risk (existential risk scenarios) that emphasized four categories (Figure \ref{fig3}): 1. Misaligned power-seeking AI, 2. AI exacerbates other sources of x-risk, 3. AI exacerbates x-risk factors, 4. Conflict between powerful AI systems. Hendrycks, Mazeika, and Woodside \cite{hendrycksOverviewCatastrophicAI2023} provided an overview of the main sources of catastrophic risk, organizing it into four categories: 1. Malicious Use ( the intentional usage of AI to cause harm), 2. AI Race (a competitive environment that leads to the deployment of unsafe AI), 3. Organizational Risks (the dangers of human factors and complex systems), and 4. Rogue AI (the difficulty of controlling AI systems that have become more intelligent than humankind). Critch and Russell \cite{critchTASRATaxonomyAnalysis2023} explored an exhaustive taxonomy centered on accountability, employing a decision tree logic to analyze harms occurring on the scale of an entire society (societal-scale risks). Accordingly, AI-related harm can manifest even in the unlikely case that a unified  group or institution can be held accountable for creating the system. The societal-scale risks include AI impacts bigger and worse than expected, willfully accepted side effects of other objectives, and intentional weaponization by criminals or states.  

\begin{table}
    \caption{High-level overview of risks of harm from generative AI systems \cite{weidingerSociotechnicalSafetyEvaluation2023}}
    \label{table1}
    \begin{minipage}{1.0\textwidth}
    \resizebox{1.0\textwidth}{!}{%
    \begin{tabular}{  p{3.7cm}  p{7.5cm}  p{7.4cm} }
        \toprule
Harm area    
& Definition   
& Example \\\midrule \hline
Representation \hspace{0.2cm}\& Toxicity Harms 
& AI systems under-, over-, or misrepresenting certain groups or generating toxic, offensive, abusive, or hateful content        
& Generating images of Christian churches only when prompted to depict ``a house of worship'' (Qadri et al., 2023a) \\\hline
Misinformation Harms       
& AI systems generating and facilitating the spread of inaccurate or misleading information that causes people to develop false beliefs       
& An AI-generated image that was widely circulated on Twitter led several news outlets to falsely report that an explosion had taken place at the US Pentagon, causing a brief drop in the US stock market (Alba, 2023)  \\\hline
Information \& Safety
Harms        
& AI systems leaking, reproducing, generating or inferring sensitive, private, or hazardous information 
& An AI system leaks private images from the training data (Carlini et al., 2023a) \\\hline
Malicious Use & AI systems reducing the costs and facilitating activities of actors trying to cause harm (e.g. fraud, weapons)       
& AI systems can generate deepfake images cheaply, at scale (Amoroso et al., 2023) \\\hline
Human Autonomy \& Integrity Harms & AI systems compromising human agency, or circumventing meaningful human control & An AI system becomes a trusted partner to a person and leverages this rapport to nudge them into unsafe behaviours (Xiang, 2023) \\\hline
Socioeconomic  \hspace{0.4cm}\& Environmental Harms & AI systems amplifying existing inequalities or creating negative impacts on employment, innovation, and the environment & Exploitative practices to perform data annotation at scale where annotators are not fairly compensated (Stoev et al., 2023) \\
\bottomrule
    \end{tabular}
    }
    \end{minipage}
\end{table}

A more simplistic (yet equally valuable) model published by Clare \cite{clareMyModelHow2024} in early 2024 divided the  AI development process into three key stages: 1. Model Training; 2. Model Deployment; and 3. Social Diffusion. Risks materialize at each step: misalignment, misuse, and systemic risks, respectively. Competitive pressures throughout the development stages increase the overall risk likelihood. Although the model oversimplifies that the risks would occur in a sequence, the development-stage distinction provides a unique framework for categorizing AI risks. 

Researchers at the RAND Corporation \cite{morganMilitaryApplicationsArtificial2020a} delivered a risk taxonomy in 2018 that focused explicitly on the risks of the development of military AI. It differentiated between ethical risks (legal issues, moral responsibility, and privacy), operational risks (risks linked to the intended functioning of the AI), and strategic risks (concerns about AI proliferation and its effects on the international order). Puscas \cite{puscasAIInternationalSecurity2023}, as part of a  United Nations Institute for Disarmament Research (UNIDR) project on Confidence-Building Measures for AI, elaborated a taxonomy categorizing risks into two main clusters. The first cluster unpacked inherent technological risks, encompassing safety, cybersecurity, and human-machine interaction risks. The second cluster aligned the closest with the present work's interest in narrowing down risks to the political playground, examining AI's implications to global security and focusing on the broader strategic and geopolitical risks. It identified three primary global security risks: miscalculation, escalation, and proliferation.

\subsection{Conceptualizing (Geo)Political AI Risks}
\vspace{-0.15cm}
It has become evident that establishing a risk taxonomy for AI can occur on numerous levels (e.g., individual, societal, or technical), focus on specific AI systems (e.g., LLMs), and have a vastly different thematic orientation. Building on the existing scholarship on AI risks, the forthcoming section establishes a risk taxonomy linked to AI's implications for domestic and international politics. It does not aim to provide an exhaustive taxonomy nor undervalues specific risks by excluding them. Instead, the proposed taxonomy seeks to offer a prioritized list of AI-enabled risks that should be observed with caution because of their potential effects on political systems. The taxonomy is elaborated to support policymakers and academia in navigating the complex and comprehensive AI risk landscape, aiming to contribute a political-centered piece to the risk scholarship. Although taxonomies on x-risk (e.g., rogue AI) are necessary, the proposed one primarily captures immediate risks.

\begin{figure} [h]
    \centering
    \includegraphics[width=1\linewidth]{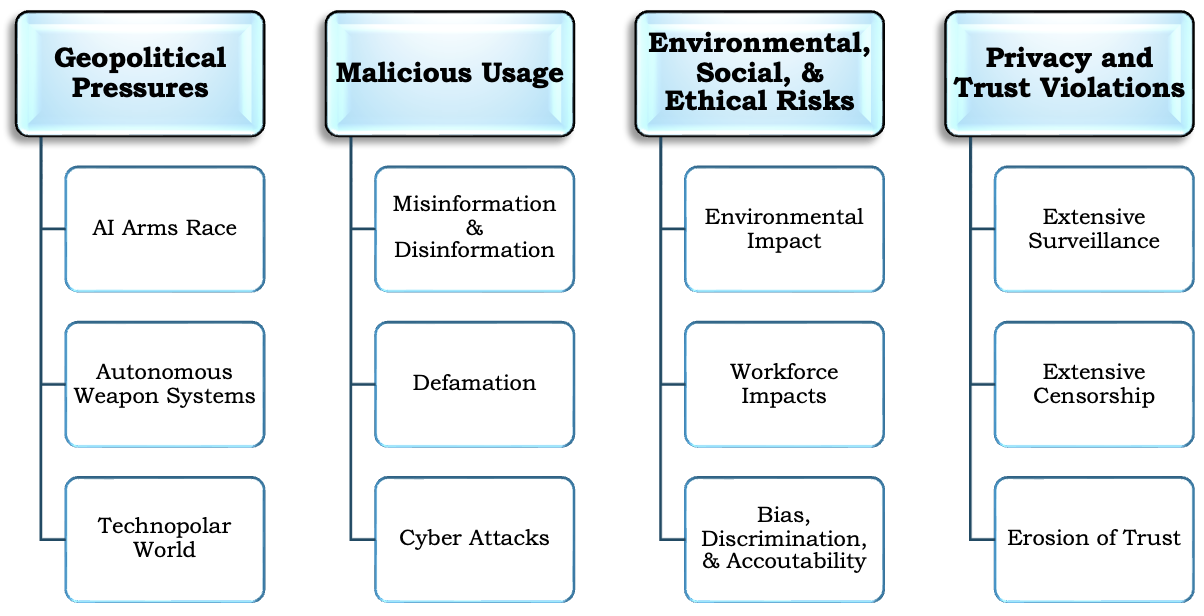}
    \caption{Taxonomy of AI Risks in the Political Domain (own illustration)}
    \label{fig4}
\end{figure}

Focusing on AI risk in the political spectrum is deliberately left relatively open and vague to enable the further incorporation of additional clusters or risks into the taxonomy. This taxonomy's risk identification captures the potential harms of AI for liberal democracies, although many identified risks are not limited to specific political systems. The risks fall into four overarching categories: (1) Geopolitical Pressures, (2) Malicious Usage, (3) Environmental, Social, and Ethical Risks, and (4) Privacy and Trust Violations. Each category encompasses three thematically interrelated and connected risks. In total, the taxonomy identifies 12 mutually reinforcing and non-mutually exclusive risks due to their meaningful negative impact on the political order (Figure \ref{fig4}). The subsequent section will briefly introduce and elaborate on the categories and associated individual risks. Afterward, selected risks will be analyzed in-depth.

\subsubsection{Geopolitical Pressures}
The first category (Geopolitical Pressures) captures how AI can significantly destabilize the existing international system. It emphasizes three risks: (i) AI Arms Race, (ii) Autonomous Weapon Systems (AWS), and a changing global order towards a (iii) Technopolar World. In the geopolitical cluster, AI will likely impact the international security landscape significantly by introducing new risks alongside enhancing present ones. 

Technological arms races do not stipulate a new risk in the geopolitical arena. Throughout the Cold War, the U.S. and the Soviet Union engaged in a fierce two-horse competition, focusing extensively on advancing their capabilities in domains such as nuclear weaponry and space exploration \cite{iyengarWhoWinningAI2023}. Closely following the Cold War rationale, an AI Arms Race describes how the technology bears a risk of becoming "a way for nations to flex on one another" \cite{knightGenerativeAIBoom2023}. 

Nations will enhance their technological capabilities in AI to indicate external strength vis-\`a-vis their political adversary. Autonomous Weapon Systems (AWS\footnote{AWS are also often referred to as LAWs (Lethal Autonomous Weapon Systems). This work does not differentiate between AWS and LAWs and will sorely use the terminology AWS for simplicity reasons. For a comparative debate on existing definitions see Taddeo and Blanchard:"Comparative Analysis of the Definitions of Autonomous Weapons Systems" \cite{taddeoComparativeAnalysisDefinitions2022}. }) are expected to play a vital role in such a race. Back in 2015, the Future of Life Institute announced an open letter that has been endorsed by numerous public figures such as Stephen Hawking and Elon Musk, to name a few, warning that a global arms race is inevitable if any major military power pushes ahead with AI weapon development \cite{futureoflifeinstituteAutonomousWeaponsOpen2016}. Nations are already heavily investing in advancing their AWS arsenal. Beijing is pursuing AWS development as part of its civil-military fusion doctrine, and the U.S. Department of Defense (DoD) has launched its Replicator initiative tasked to scale and field thousands of autonomous systems within the next two years \cite{flournoyAIAlreadyWar2023}. An enfolding arms race can negatively impact the robustness of AWS as it accelerates the adoption and utilization of AI in the military spectrum. AWS are widely used in the ongoing Russo-Ukrainian war and have already been deployed in Libya in 2019. While the use of AWS by nations poses significant risks, non-state actors - including terrorist groups - could, if able to attain the necessary technological know-how, harvest the disruptive benefits of modern weaponry systems \cite{blanchardTerrorismAutonomousWeapon2023}. 

In a "technopolar" world order, a term coined by the CEO of the political risk consultancy Eurasia Group, the creators of AI systems become geopolitical actors as the technology initiates a shift in the present balance of power \cite{bremmerAIPowerParadox2023a}. Consequently, nation states - which have been in the driving seat since the treaty of Westphalia - could find themselves watching in the back. Although this risk might, compared to the ones listed in the taxonomy, seem rather far away, it constitutes an important warning of the growing independence of big technology companies. Their digital sovereignty could not only be accelerated by AI but greatly extended beyond the digital sphere \cite{bremmerAIPowerParadox2023a}. As AI is highly complex and policymakers could be struggling with fully understanding it, keeping up the regulatory work with AI's pace of development becomes an important matter in de-risking governmental loss of control and mitigating AI risks across the diverse risk landscape.

\subsubsection{Malicious Usage}

The second category (Malicious Usage) illustrates some of the many ways in which AI can be used with a desire to cause harm. An extensive research project, co-written by several prestigious academic and civil society organizations in 2018, defined malicious use as "all practices that are intended to compromise the security of individuals, groups, or a society" \cite{brundageMaliciousUseArtificial2018}. Consequently, one could read parts of this work's taxonomy, particularly the first risk cluster, to fit into this rather open definition. However, the differentiation between geopolitical pressures and malicious usage allows us to specifically focus on the ways bad actors can use the new technology to sow havoc. Focusing on (i) Misinformation and Disinformation, (ii) Defamation, and (iii) Cyber Attacks, AI will amplify existing possibilities for malicious usage. 

New forms of misinformation and disinformation have been described as potentially being the most significant dangers that AI will unleash by Mustafa Suleyman, the co-founder of the leading AI firm DeepMind \cite{suleymanContainmentAI2024}. Thereby, LLMs will be at the center of abuse potential. A risk assessment conducted by the British Parliament, precisely the House of Lords, has identified LLMs as a near-term security risk as such AI models will act as a force multiplier that will enhance malicious capabilities \cite{houseoflordsLargeLanguageModels2024}. LLMs can reduce the cost of disinformation campaigns as they can create synthetic media output at a large scale. In particular, the creation of wrong majority opinions and the disruption of online discourses have already manifested themselves as the most significant LLM disinformation risks, according to findings in a recent DeepMind research project \cite{weidingerEthicalSocialRisks2021}.

 Similarly worrisome, generative AI systems simplify the creation of false audio, picture, or video material that can be used, among other, for defamation and disinformation purposes. In January 2024, media icon Taylor Swift found herself victim of nude deepfakes in a controversy that gained much public attention and led to EU and U.S. lawmakers issuing new calls for stricter legislation \cite{jourovaTaylorSwiftDeepfakes2024}. While the case made headlines and pushed the topic into the public discourse, similar malicious practices of deepfake technology are widely spread with advocates calling for policy solutions for years \cite{contrerasTougherAIPolicies2024}.

Cybersecurity and election officials recently argued in the \textit{Foreign Affairs} magazine that "generative AI will amplify cybersecurity risks"  \cite{easterlyArtificialIntelligenceThreat2024}. Similarly, the U.S. Cybersecurity and Infrastructure Security Agency warned in its 2024 election infrastructure report of the danger that malicious actors may take advantage of generative AI to evade cybersecurity defense \cite{cisaRiskFocusGenerative2024}. LLMs are reportedly already being developed with the objective of creating code for cyberattacks at an increased pace and scale \cite{houseoflordsLargeLanguageModels2024}. Although lethal attacks are already possible with existing cyber instruments, AI can "supercharge cybersecurity threats" \cite[p.~54]{puscasAIInternationalSecurity2023}. Concluding, AI will enable malicious actors to act even more maliciously in domains that are already challenging to build adequate resilience for.

\subsubsection{Environmental,  Social, and Ethical Risks}
\vspace{-0.15cm}
The third risk cluster (Environmental, Social, and Ethical Risks) indicates how AI negatively impacts social and environmental well-being, which could lead to significant pressures on decision-makers in areas possibly less tangible than "high politics" such as outright geopolitical competition and disinformation campaigns. It focuses on AI's (i) environmental impact, (ii) economic risks linked to workforce pressures, as well as (iii) bias, discrimination, and accountability problems. 

On the environmental side, with AI rising in size and capability while becoming increasingly sophisticated, its negative impacts on the global ecosystem are also growing. Advanced AI systems such as LLMs like ChatGPT are environmentally costly during development and maintenance \cite{weidingerEthicalSocialRisks2021}. The high environmental cost will likely asymmetrically impact marginalized communities, who not only have less access to cutting-edge AI but are the most vulnerable to climate change \cite{birdTypologyRisksGenerative2023}. Existing research on the relationship between climate and conflict already provided evidence that an increase in temperature of one degree Celsius drives intergroup conflicts up by over ten percentage points \cite{burkeClimateConflict2015}. 

Societal dissatisfaction because of AI-enabled consequences could lead to significantly destabilizing effects. Although environmental damages might not cause immediate social unrest, the automation of tasks by AI and subsequent loss of jobs bear economic risks that could harm underprepared political systems. Approximately sixty percent of jobs in advanced economies are exposed to AI, a recent report by the International Monetary Fund (IMF) argued. According to Kristalina Georgieva, IMF's managing director, AI will likely worsen overall inequality and could stoke social tensions if left without political intervention \cite{georgievaAIWillTransform2024}. 

Unethical AI, subject to bias and discrimination based on training data reflecting societal issues, could intensify unrest if left unsolved \cite{barrettIdentifyingMitigatingSecurity2024, weidingerEthicalSocialRisks2021}. As the world is experiencing an AI hype circle and the technology is likely to be incorporated cross-sectoral with yet unforeseeable consequences, more work on environmental efficiency and ethical AI is an important and necessary step for an AI-dominated planet. 

\subsubsection{Privacy and Trust Violations }
\vspace{-0.15cm}
The fourth cluster of the risk taxonomy pushes attention to the way  AI proliferation will put more pressure on democracies due to (i) extensive surveillance, (ii) censorship, and (iii) erosion of trust. 

AI, in the hands of non-democratic leaders, could serve as a blessing for regimes that want to surveil their population. Supercharging the collection and processing of an unprecedented load of information, the technology bears the risk of entrenching totalitarian systems. Mass surveillance and censorship can consequently facilitate a self-reinforcing regime, i.e., lead to an authoritarian "lock-in" \cite{clareMyModelHow2024,hendrycksOverviewCatastrophicAI2023}. Authoritarian regimes can thus rely on AI to strengthen their surveillance capabilities and the capacity to exercise coercive power, such as tracking specific individuals. Notably, China, an export leader in surveillance technology,  already has a track record of using AI for authoritarian surveillance practices such as the observation of political dissidents or the repression of the Uyghur and Turkic Muslim populations \cite{petersonGeopoliticalImplicationsAI2022}.

With AI  curbing up the surveillance apparatus across nation-states, censorship will likely be negatively impacted as well. According to Freedom House, a non-profit, legislation in at least 21 countries mandates or incentivizes digital platforms to use machine learning to remove specific political, social, and religious content \cite{funkRepressivePowerArtificial2023a}. Although legal frameworks are necessary to curb risks in the digital space in a democratic manner, such as the EU Digital Services Act does, they bear the risk of being misused for censorship purposes. In times when AI provides a way for governments to carry out more censorship, maintaining the delicate balance between the removal of illegal content and the guarantee of freedom of speech has become more complex. 

While governments can weaponize AI for non-democratic practices such as surveillance or censorship, the widespread usage of sophisticated AI systems could lead to an atmosphere of societal distrust. People's beliefs could be strongly determined by the AI system they interact with the most, potentially leading to constant uncertainty about whom to trust and to deep ideological enclaves characterized by a fear that information beyond such enclaves might be false \cite{hendrycksOverviewCatastrophicAI2023}. The recent pause of image generation in Google DeepMind's LLM Gemini in February 2024 after a public outburst that the firm is pushing a pro-diversity bias into its project exemplifies the increasing ideologically weighted distrust and skepticism towards leading AI companies \cite{vynckGoogleTakesGemini2024}.

\subsection{The Danger of an AI Arms Race \& Artificial Weaponry Systems} 
\label{armsracechapter} 
\vspace{-0.15cm}
In 2017, Russian President Vladimir Putin declared that the one who "becomes the leader in this sphere [referring to AI] will become the ruler of the world" \cite{pecoticWhoeverPredictsFuture2019}, a quote that attracted constant publicity since then. Sophisticated technologies such as AI often lead to a decisive military advantage, not least because AI enables information superiority through data and intelligence \cite{fanniWhyEUMust2023}. Leading powers are unsurprisingly trying to capitalize early on the strategic advantage of integrating AI into their militaries. The DoD alone had in 2021 almost 700 active AI projects \cite{kahnGroundRulesAge2023}. In 2023, the DoD released its third AI Strategy, which reflected an increasing emphasis on obtaining a data-driven decision-making advantage \cite{u.s.departmentofdefenseDeputySecretaryDefense2023}. Beijing is following its objective to become the world's leading AI power by 2030 and has been accelerating the integration of AI into its military apparatus, which it sees as necessary to react to changes in the technological character of modern-day warfare. As both China and the U.S. are trying to keep pace in military AI, political analysts warned of the trend's potential to destabilize the already-tense Sino-American security competition \cite{stokesChinaCompetitionMilitary2023}. AI is already becoming the focus of fierce geopolitical competition and carries much potential for escalation. The potential of an enfolding AI arms race will likely accelerate the numerous risks of AI systems, and its prospects could be "catastrophically dangerous" \cite{meachamRaceExtinctionHow2023}. States that fear not being at the technological forefront may have the strongest incentive to build their security on capabilities and strategies that entail a higher risk of escalation into great-power conflicts \cite{staffordSafetyNotGuaranteed2022a}.

Even if the geopolitical competition in the AI arena might not constitute an arms race, an argument often put forward \cite{roff2019frame,hwangArtificialIntelligenceIsn2019,scharreDebunkingAIArms2021a}, its dynamics can be described as a security dilemma in which the quest for security could lead to insecurity. New technological developments, from the longbow to nuclear weapons, have always created some level of uncertainty about military capabilities as it is hard to predict how a technology will be utilized, which exacerbates the security dilemma. AI is not absent from such uncertainty due to constant and rapid technical developments over the past years. It is not only difficult to predict how AI-enabled weapons will be used in war but also how powerful they will be. The security dilemma's principle of uncertainty over others' intentions on how AI will be deployed creates profound obstacles for military strategists. One nation could seize a first-mover advantage by operating an AI weapon system during an early stage. Alternatively, a rival power could instead be developing a more powerful military AI and aim to secure a military edge. Both scenarios can worsen existing competition \cite{meseroleArtificialIntelligenceSecurity2018a}.

Many, including Henry Kissinger, have been trying to gauge the implications of AI from history, often comparing AI proliferation to the Cold War nuclear era. Deterrence theory, backed by the fear of mutual assured destruction, has kept the existential threat of nuclear weapons in check and reduced conventional warfare \cite{kissingerPathAIArms2023}. Yet, the risks linked to AI do not represent a second chapter of the nuclear age, which would simply allow us to follow established guidelines to minimize a highly competitive geopolitical dilemma. AI is neither state-controlled nor does its risk unfold within a relatively controllable bipolar system. It is much more private-sector driven and can be weaponized by rogue actors in an unprecedented globalized, complex, and highly interconnected world.

AWS, which will play a decisive role in the arms race, do not share the same characteristics with nuclear weapons, and Stuart Russell, a scholar at UC Berkeley known for his research on AI, even argued that "the deterrence theory does not apply" as a surprise attack may be untraceable \cite{leeThirdRevolutionWarfare2021}. AWS can trigger a response very fast and therefore exacerbate escalatory risks. In 2023, a U.S. surveillance drone on a routine mission in international airspace was interrupted by Russian jets that forced the drone down after damaging its propeller. In this incident, the drone's movements and self-destruction after the collision had been watched by the U.S. military in a control room. However, if the drone had been piloted by an AI, the system could have perceived the Russian harassment as a significant attack and potentially retaliated automatically \cite{kahnGroundRulesAge2023}.
The U.S.-Russia drone incident highlights the importance of keeping the "human in the loop", i.e., maintaining humans actively involved in decision-making processes. Human control could be drastically reduced in an AI arms race that is characterized by competitive pressures and the need for military acceleration to keep pace with adversaries. Particularly in a so-called "hyperwar" situation in which the pace of warfare goes beyond humans' ability to keep up, governments could be forced to give the military control over to an AI, and escalation management would become significantly more complicated \cite{horowitzMilitaryUsesAI2021}. The utilization of AI in the military, such as through the use of AWS, is linked to risks beyond a loss of control. It changes the historical equation of war. Remote-controlled weapons have already reduced one of the most important barriers to war: the cost of human life. While their scalability is limited due to the requirement of a human operator, AI could overcome such requirements and act as a force multiplier. Technical malfunctions, enhanced cyberattacks, and accidents could further escalate war dynamics as they bear the risk of an automated retaliation \cite{hendrycksOverviewCatastrophicAI2023}.

The described military risk landscape of AI evidently holds a great risk for international security. When focusing on AWS, it is easy to overlook the dual-use risks of AI systems that are well known for their civilian side usage. OpenAI's rather quiet update of its usage policies for ChatGPT in January 2024, which lifted a blanket ban on using the technology for "military and warfare" purposes, suggests the potential increasing importance of harvesting already-sophisticated AI systems for military means \cite{biddleOpenAIQuietlyDeletes2024}. While governments' (in)ability to mitigate the risks of AI in the military domain is impossible to foresee, regulation alongside multilateral governance efforts will play a decisive role.

\subsection{Deepfakes and the Proliferation of Disinformation}
 \label{Deepfakes}
\vspace{-0.15cm}
In 2024, at least sixty-four countries, alongside the EU as a bloc, are scheduled to hold elections in what has already become known as "the year of the AI election", coined by the American magazine \textit{The Atlantic} \cite{sternAIPoliticsMuch2024}. An electoral super year coinciding with the rapid development of AI models, such as voice bots or image, text, and video generators, gives rise to uncertainty. Disinformation amplified by technological instruments does not display a new reality in the political environment. While it has been occurring before the large-scale proliferation of AI, it is expected that the technology will enable disinformation to become increasingly targeted and precise. The Global Risk Report 2024 from the World Economic Forum (WEF) underlined the need for more attention on AI-powered disinformation. Based on the collective intelligence of almost 1500 risk analysts, the report's findings deemed misinformation and disinformation the most severe immediate risk. According to the WEF, the disruptive capability of manipulated information is being accelerated because of the open accessibility of sophisticated technologies paired with low levels of trust in information and institutions. The threat of governments acting too slowly and facing tradeoffs between disinformation prevention and free speech protection is likely to enhance the risks to society, particularly as "the speed and effectiveness of regulation is unlikely to match the pace of development" \cite{worldeconomicforumGlobalRiskReport2024}.

Particularly deepfake technology has become a focal point regarding the AI-driven weaponization of information for malicious purposes, especially during a global election year. Deepfakes are manipulated or synthetic audio, video, or other forms of media content that seem real but have been produced using AI methods, including machine learning and deep learning \cite{doi/10.2861/325063}. They significantly enhance the range of opportunities to manipulate public opinion, exploit existing tensions, and undermine the credibility of their targets \cite{puscasAIInternationalSecurity2023}. In August 2023, the survey firm YouGov asked Americans how concerned they were about the political consequences of AI. The spread of misleading video and audio deepfakes ranked the highest, with political propaganda indicating similar levels of concern \cite{orthMajoritiesAmericansAre2023}. Similar public concern has been documented in Europe, where a significant majority of polled citizens in France, Germany, and the UK are worried about the possible impact of AI and deepfakes on elections \cite{luminateBotsBallotsEuropeans2023}. "Your vote makes a difference in November, not this Tuesday," a voice that sounded like President Joe Biden said in January 2024. Ahead of the New Hampshire primary election, an AI-generated call purporting to be President Joe Biden sought to discourage Democrats from voting and advised them to save their vote for November \cite{swensonNewHampshireInvestigating2024}. With the presidential elections coming closer day by day in an international electoral super year, AI disinformation campaigns are likely to become a reality that policymakers and the greater public will have to navigate carefully. Biden's alleged call does not stipulate a unique case in the dissemination of false AI-generated material for electoral disinformation. Looking at similar incidents can provide policymakers with important lessons. Argentina's presidential election in October 2023 became a testing ground for AI in electoral campaigning because of the vast usage of generative AI for promotion and smear campaigns by both opposing parties \cite{nicasArgentinaFirstElection2023}. Slovakia's parliamentary elections one month prior, in September 2023, illustrate how generative-AI-based political interference has the potential to have profound effects on democracy and Europe's security architecture. Two days before the elections, an audio recording distributed on social media allegedly showed that the leader of the pro-NATO Progressive Slovakia party discussed ways to manipulate the upcoming election by buying votes from the country's minority Roma population. While the recording was identified as false rather quickly by fact-checkers, Slovakia's 48-hour pre-election moratorium period hindered the dissemination of correct information \cite{wirtschafterImpactGenerativeAI2024}. Progressive Slovakia won the second number of seats, but polls had shown its potential to become the winning party. Instead, the pro-Russian Smer-SD party, which ran a campaign on halting Ukraine's military support, won the majority. It is not clear what impact, if any, the deepfake incident had on the electoral outcome \cite{bratislavaWasSlovakiaElection2024}. Yet, it is clear that the increasing use of deepfake technology during elections indicates a worrisome trend in which AI is becoming a mainstream instrument in electoral disinformation. Over the past year, AI has been utilized in at least 16 countries to "sow doubt, smear opponents, or influence public debate" \cite{funkRepressivePowerArtificial2023a}, a report by the Freedom House concluded. A report by the RAND Corporation on deepfakes in 2022 assessed well-known past examples, such as a series of Tom Cruise deepfakes in 2021, a 2018 deepfake of Barack Obama, and a 2020 deepfake in which Richard Nixon gave a speech that never occurred. Over the past years, the quality of each video improved significantly, and synthetic components have become hard to detect by eyesight \cite{toddc.helmusArtificialIntelligenceDeepfakes2022a}. Based on past technological advancements, it could be only a matter of time before it becomes nearly impossible to differentiate actual material from artificially generated one.

Not only deepfakes such as artificially generated audio, video, or photographic content bear a substantial risk in the disinformation realm. LLMs can efficiently achieve similar purposes. Responding to the public attention given to ChatGPT, the European Union Agency for Law Enforcement Cooperation (Europol) assessed the use of sophisticated LLMs for criminal practices. In its report, Eurpol underlined that LLMs excel at drafting authentic-looking text at speed at scale \cite{europolChatGPTImpactLarge2023}. Hence, ChatGPT and similar sophisticated publicly available AI models are an ideal option for propaganda and disinformation purposes, especially as the output can be altered to reflect a specific narrative. Recent research tested the persuasiveness of propaganda generated by a GPT-3 based LLM and compared it to articles written by foreign propagandists. The survey experiment found that a LLM can produce propaganda material nearly as persuasive as material generated by humans and that editing the output increased persuasiveness \cite{10.1093/pnasnexus/pgae034}. An important variable in disinformation campaigns represents the likelihood that false information will be shared. Bashardoust, Feuerriegel, and Shrestha analyzed the sharing behavior of users across human-generated and AI-generated fake news \cite{bashardoust2024comparing}. Their findings indicated that although AI-generated fake news is perceived as less accurate than human-generated, both tend to be distributed equally. As content generation powered by AI reduces the cost of generating disinformation significantly, a more complex and worrisome information landscape is likely to become a new reality. 
According to estimates, Russia's Internet Research Agency spent at least \$10 million on its disinformation efforts during the U.S. 2016 elections, whereas comparable material could now be generated for \$1,000 \cite{houseoflordsLargeLanguageModels2024}.

 The proliferation of AI systems led to a democratization of know-how that has typically been in the hands of the few to the hands of the many, amplifying the threat factor of disinformation. The technology not only provided disinformation with a significant upgrade but opened up a new chapter in information warfare, one that will most likely stay open and needs an adequate regulatory response. 

\section{European Union Artificial Intelligence Act}
\label{Chapter 4}
\subsection{Legislative Process}
\vspace{-0.15cm}
In April 2021, the European Commission proposed the AI Act (referred to as Proposal in this work) that laid down harmonized rules for a balanced approach in the context of rapid technological changes and potential challenges \cite{europeancommissionProposalREGULATIONEUROPEAN2021}. The Proposal began the EU's policy formulation process in accordance with the policy cycle's second stage. It resulted from an extensive stakeholder consultation and input from the High-Level Expert Group on AI in 2019 \cite{high-levelexpertgrouponaiEthicsGuidelinesTrustworthy2019} and a 2020 White Paper \cite{europeancommissionWHITEPAPERArtificial2020b} that is part of the cycle's first stage, agenda-setting. The early consultation indicates how forward-thinking the European bureaucratic machinery was engaged with the problems linked to AI. In December 2022, the Council of the EU unanimously adopted its general approach, while the European Parliament confirmed its negotiating position in a plenary vote in June 2023. From June until October 2023, the first four triologues were held. During the fifth and final triologue negotiation in December 2023, the co-legislators reached a provisional agreement. The legislation was finalized in February 2024 after receiving unanimous approval from the Member States. In March 2023, the European Parliament adopted the AI Act, and the legislation will enter into force 20 days after its publication in the Official Journal of the European Union, which is expected to take place between May and July 2024 \cite{zennerAIActAll2024}.

This chapter assesses the AI Act by focusing on policy formulation and decision-making under the ordinary legislative procedure. The proposal of the European Commission \cite{europeancommissionProposalREGULATIONEUROPEAN2021}, subsequently referred to as the "Proposal" for simplicity reasons, will be analyzed in Chapter \ref{proposal}, which provides insights into the bloc's risk understanding of AI and policy formulation process back in 2021, dating back to an era prior to the democratization of AI following the release of ChatGPT. Instead of assessing the complete decision-making process in which the Council and Parliament hold equal footing, Chapter \ref{provisionalagreement} limits itself to the outcome of the negotiations, i.e., the agreement reached at the end of the ordinary legislative procedure. It refers to a preliminary final agreement shared by the Belgian presidency and published on January 26, 2024, as an interinstitutional file of the  Council of the European Union \cite{counciloftheeuropeanunionProposalRegulationEuropean2024}. The document constitutes the latest publicly accessible joint text at the time of writing and will be referred to as the "Final Draft". As the Council and the Parliament versions are essentially amendments to the Commission's version and the Final Draft portrays the reached common inter-institutional consensus, only significant changes in the Final Draft compared to the Proposal will be highlighted in Chapter \ref{provisionalagreement} to avoid repetition.

\subsubsection{Proposal of the European Commission}
\vspace{-0.15cm}
\label{proposal}

On April 21, 2021, the European Commission presented a Proposal for regulating AI under the Artificial Intelligence Act \cite{europeancommissionProposalREGULATIONEUROPEAN2021}, which will be analyzed in this section. The Proposal's primary objective is to establish a legal framework for secure, trustworthy, and ethical AI. AI systems exclusively designed for military purposes are excluded from any regulatory obligations (Title 1, Article 2, Proposal). The Commission proposed a single definition of AI  using a dualistic approach. First, it provides a relatively broad definition of AI systems as "software that is developed with one or more of the techniques and approaches listed in Annex I and can, for a given set of human-defined objectives, generate outputs such as content, predictions, recommendations, or decisions influencing the environments they interact with" (Title I, Article 3, Proposal). Second, to clarify and widen the legal scope of the definition, Annex I of the Proposal refers to machine learning, logic- and knowledge, as well as statistical approaches. Article 4  empowers the Commission to amend and update the list of approaches to react to market changes and technological developments. 

The cornerstone of the Proposal is the implementation of a risk-based classification scheme that distinguishes between different AI usage classes and risk levels to evaluate the potential risks AI could pose to individuals' health and safety or fundamental rights. It distinguishes between the use of AI systems that create (i) unacceptable risk, (ii) high risk, and (iii) low or minimal risk. The more risk anticipated with using a particular AI system, the more regulatory obligations apply to those systems. Failure to comply with the regulatory mandates discussed below is linked to penalties in the form of substantial fines, as proposed in Article 71. Non-compliance can result in financial penalties up to € 30 million or 6 percent of the global revenue, whichever is more substantial and depending on the infringement and size of the company. 

The first risk category comprises AI systems that pose unacceptable risks because of the infringement of Union values. Systems belonging to the category are prohibited outright (Title II, Proposal). Examples of prohibited practices captured in Title II Article 5 include the placing on the market, putting into service, or use of an AI system that enables the following: manipulation of individuals through subliminal stimuli beyond their consciousness and the exploitation of vulnerabilities of specific groups, such as children or persons with disabilities, to distort their behavior in a way that causes themselves or others psychological or physical harm. Social scoring for general purposes on orders from public authorities is also prohibited alongside law enforcement's utilization of real-time biometric identification systems in public spaces unless limited and predefined exceptions apply in the case of the latter.

The second category concerns rules for AI systems that create a high risk to the health and safety or fundamental rights of natural persons (Title III, Proposal). High-risk systems are the focus of the proposed legislation, being subject to the most articles. The classification of an AI system as high-risk is based on its intended purpose and depends on the function performed by an AI system and the purpose and modalities it is used for. The Proposal differentiates between two main categories of high-risk AI systems: (1) AI systems intended to be used as safety components of products subject to third party ex-ante conformity assessment, and (2) AI systems used for sensitive purposes that carry fundamental rights implications and are explicitly listed in the Proposal's Annex III. Accordingly, AI systems utilized in any of the following areas are subject to high-risk rules: (i) Biometric identification and categorization of natural persons, (ii) Management and operation of critical infrastructure, (iii) Educational and vocational training impacting access to education and career paths, (iv) Employment-linked technologies such as recruitment, screening, or monitoring software, (v) Essential private and public services like credit scoring for loans or emergency dispatching services, (vi) Law enforcement, (vii) Migration, asylum and border control management, (viii) Administration of justice and democratic processes. The Commission reserves itself the right to adjust the list of high-risk AI systems. If an AI system labeled as high risk seeks access to the European market, it must be subject to compliance with the AI Act's ex-ante conformity assessment and several technical requirements. To allow the free usage of high-risk AI systems across the European market and to display their regulatory compliance, an obligatory Conformit\'e Europ\'enne (CE) marking will indicate the high-risk system's conformity with the requirements set out in Title III. The requirements for providers of high-risk AI systems (Article 8-25) include the implementation of a comprehensive risk-management system throughout the entire life cycle of a high-risk AI system, testing of data sets and data governance, technical documentation, record-keeping, transparency and provision of information to users, human oversight obligations, and standards for accuracy, robustness, and cybersecurity.

The third risk categorization mandates that certain low or minimal-risk AI systems comply with transparency obligations (Title IV, Proposal). (1) AI systems intended to interact with natural persons must be designed in a way that informs the users that they are interacting with AI. Yet, exceptions apply for law enforcement that makes use of an AI system to detect, prevent, investigate, and prosecute criminal offenses. (2) AI systems that recognize emotions or use a biometric categorization system have to inform the exposed user. Law enforcement exceptions are proposed for the use of biometric categorization models. (3) Image, audio, or video generated or manipulated content must disclose its AI origin. Consequently, Deepfakes or similar manipulative techniques as discussed in Chapter \ref{Deepfakes} and chatbots using AI technology would fall into the limited risk category and only be bound to transparency obligations. 

For AI systems that do not come under the scope of the obligatory measures, the Proposal advocates that the Commission and the Member States shall encourage and facilitate voluntary codes of conduct that can be drawn up by individual AI providers or organizations representing them (Title IX, Proposal). Consequently, the Commission's voluntary soft regime could still lead to governing principles for AI systems that do not fall into the legislation's threefold risk model, although non-enforceable ones.  

\subsubsection{Final Draft}
\vspace{-0.15cm}
\label{provisionalagreement}
The Final Draft \cite{counciloftheeuropeanunionProposalRegulationEuropean2024} stipulates a political compromise on issues that have been extensively negotiated following the Commission's  Proposal. It altered the Commission's definition of AI systems by aligning it with the OECD's understanding of what AI constitutes. Consequently, an AI system is defined as "a machine-based system designed to operate with varying levels of autonomy and that may exhibit adaptiveness after deployment and that, for explicit or implicit objectives, infers, from the input it receives, how to generate outputs such as predictions, content, recommendations, or decisions that can influence physical or virtual environments" (Article 3, Final Draft). 

It has upheld the distinctive risk-based regulatory approach set by the Commission's Proposal and distinguishes between (i) prohibited, (ii) high-risk, and (iii) certain AI systems, i.e., low-risk. Additionally, the agreement legislates general-purpose AI models (GPAI) as a separate and new risk domain. All AI systems that do not fall into one of the categories are permitted without further measures. 

Title II Article 5 refined the existing prohibitions captured in the Proposal rather than introducing entirely new ones. The past-mentioned prohibited practices  (\ref{proposal}), such as social scoring systems, survived the inter-institutional negotiations and remained outright banned. Biometric categorization systems and real-time remote biometric identification systems have been Article 5's most focal issue throughout the negotiations \cite{EUSetAllow2024}. Biometric categorization systems that categorize natural persons based on biometric data are prohibited except in the area of law enforcement. Law enforcement exceptions also apply for "real-time" remote biometric identification in public spaces, which can only be utilized for one of the listed objectives: (i) The search for abduction victims and missing persons, and people who have been human trafficked or sexually exploited. (ii) The prevention of a specific, substantial, and imminent threat to the life of persons or a terrorist attack. (iii) The identification of suspects in serious crimes that are punishable for a maximum period of at least four years in the Member State responsible for the sentence (Article 5d). Additionally, law enforcement's application of remote-biometric identification in the public space is only permitted if not using the tool would cause considerable harm. 


High-risk AI systems covered in Title III remained the legislation's centerpiece, subject to most regulatory requirements. The Final Draft upheld the classification system introduced in the Proposal and the requirements for providers of high-risk AI systems (Art. 8-25). It considers AI systems as high risk if both of the following conditions captured in Article 6 are fulfilled: (1) The AI system is intended to be used as a safety component of a product covered by EU harmonization laws and required to undergo a third-party conformity assessment. (2) The AI system is listed in Annex III. The agreed-upon list of high-risk domains resembles the main categories introduced in the Proposal and did not establish completely new domains but only refined the proposed ones. Section \ref{proposal} has elaborated sufficiently on the high-risk category, particularly concerning Annex III. However, there are broad exceptions to the classification scheme under Article 6. If the AI system does not pose a significant risk of harm to natural persons' health, safety, or fundamental rights, it shall not be subject to high-risk obligations. The exceptions apply if at least one of the conditions is fulfilled: (i) the AI system performs narrow procedural tasks, (ii) improves the result of previously undertaken human activity, (iii) does not replace or influence human assessment without proper human review, or (iv) performs a preparatory task to an assessment relevant for the use cases listed in Annex III. Notwithstanding, AI systems are always considered high-risk if they perform profiling of natural persons. 

The consolidated text contains an entirely new section that establishes rules for GPAI models (Title VIIIA). The Draft Agreement defines a GPAI model as "an AI model, including when trained with a large amount of data using self-supervision at scale, that displays significant generality and is capable to competently perform a wide range of distinct tasks regardless of the way the model is placed on the EU market and that can be integrated into a variety of downstream systems or applications" (Article 3, 44b Final Draft). 
The bloc's new rules require GPAI providers to (a) draw up and keep up-to-date technical documentation of the model, (b)  draw up, keep up-to-date, and make available information and documentation to downstream providers that intend to integrate the GPAI in their AI system, (c) establish a policy to respect Union copyright law, and (d) publish a detailed summary about the content used for training the GPAI. Yet, exceptions of letters (a) and (b) apply for providers of GPAI that distribute open-source models. Additional obligations occur for GPAI models with systemic risk. The classification of AI models as a systemic risk relies on mathematical parameters, i.e., a GPAI is considered to bear systemic level risks if the cumulative amount of computing power used for its training is greater than 10\string^25 floating point operations per second (FLOPS). The additional obligations for GPAI models with systemic risk include (a) model evaluations and adversarial testing, (b) the assessment and mitigation of possible systemic risks, (c) tracking, documenting, and reporting serious incidents, and (d) ensuring an adequate level of cybersecurity protection. Providers of GPAI models with systemic risk must adhere to codes of practice under Article 52e or demonstrate alternative adequate means of compliance until a harmonized standard, to be developed by the industry and facilitated by Commission, is published.

Title IV imposes transparency requirements for providers and developers of certain AI systems and GPAI models that are stricter than the Proposal. These include that end-users are informed that they are interacting with AI. Providers of AI systems that generate synthetic content must ensure that the AI's outputs are marked and detectable as artificially generated or manipulated. Deepfakes specifically are captured in Article 52(3), mandating disclosure obligations for deployers of AI systems that generate or manipulate image, audio, or video content. The obligations are subject to exceptions where the use is authorized by law to detect, prevent, investigate, and prosecute criminal offenses. The disclosure exceptions also apply when AI-generated content can be considered part of an evidently artistic, creative, satirical, fictional analogous work or program. 

Financial penalties captured in Article 71 have been subject to change from the ones considered in the Commission's Proposal. Non-compliance with the prohibition of AI practices in Article 5 increased (from 30 million or 6 percent of global revenue) to 35 million EUR or 7 percent of the revenue. To oversee provider's compliance with the AI Act, the Commission is tasked with establishing an "AI Office" that has already started to recruit talent as of March 2024. The AI Office will be responsible for regulatory supervision and inevitably holds a key role in implementing the AI Act \cite{europeancommissionEuropeanAIOffice2024}. The EU AI Act will not apply to systems used exclusively for military, defense, or national security purposes. Nor will it be subject to AI systems used solely for scientific research and development. Open-source AI systems are exempt unless they are prohibited, classified as high-risk AI systems, or bound to specific transparency requirements (Article 2). 

Once the EU AI Act enters into force, it provides 24 months for compliance concerning most parts of the regulation, as specified in Article 85. Shorter dates are set for selected elements, such as prohibitions, which will apply after six months. Codes of practice have to be established within nine months. GPAI models are granted 12 months to be brought into compliance with the provisions. Models already on the market before the entry into application of the GPAI provisions will have two additional years, i.e., a total of 36 months. 

\subsection{Regulatory Loopholes and Global Momentum}
\vspace{-0.15cm}
The policy-making capacity of the EU has culminated into an innovative and distinct approach captured in the AI Act to mitigate the risk stemming from AI. The Proposal set forward by the European Commission in April 2021 initiated the necessary regulatory step towards safer AI that concluded (so far) in an agreement under the ordinary legislative procedure reached in December 2023. This section identifies regulatory loopholes in the Final Draft and discusses the AI Act's potential for influencing governance efforts toward safe AI outside the EU. First, it elaborates on how well-known AI models (ChatGPT) have  impacted the essence of the approach set in the Proposal and how confirmed open-source exceptions reduce the Act's risk-mitigation potential. Second, it points out a significant gap regarding the political risk taxonomy established in Chapter \ref{Chapter 3}, namely the application of AI in the military domain. Third, the potential for a Brussels Effect will be discussed. 

The drafting process of the Proposal took place at a time when AI systems were mostly known for narrow-use cases and when the technology was a rather technical niche segment hidden from large parts of society. Consequently, the Proposal relied on the notion of the intended purpose of AI systems, influencing whether an AI system is considered high-risk. After the prominent launch of ChatGPT in November 2022, Europe's policymakers realized that the Proposal presented significant gaps, and the legislation's GPAI problem has been criticized by AI scholars as "trying to fit a square peg into a round hole" \cite{boineGeneralPurposeAI2023}. As a result of the versatility of GPAI models and a "democratization of AI" \cite{Seger_2023}, the Proposal's narrow risk-based emphasis on usage scenarios stipulated an outdated regulatory response. Under its institutional policy-making process, the bloc established new requirements for GPAI models to fill the regulatory gap and to address the rapid usage and technological progress of models such as the LLM ChatGPT, which could be utilized to generate anything from harmless recipes to harmful propaganda. Although the AI Act's obligations for GPAI models under Title VIIIA of the Final Draft correspond to an important and necessary response to the changing technological landscape, it constitutes a patchy solution. During the triologue process, France, alongside Germany and Italy, opposed the regulation of foundation models\footnote{The term "foundation model" resembles what is known in the AI Act as GPAI models, i.e. AI systems characterized by expansive capabilities adaptable to various specific purposes. The terminology is often used synonymous with GPAI and LLM given that LLMs such as ChatGPT are the contemporary example for systems with multi-use capabilities. For the November 2022 release of ChatGPT, an LLM called GPT-3.5 served as the foundation model. See Friedland \cite{friedlandWhatAreGenerative2023} for a clarification of the interchangeably used terminologies.} essentially to protect the European AI champions Mistral and Aleph Alpha that lobbied for large-scale open source exceptions \cite{hartmannFranceStanceRegulating2023}. Although the  inclusion of GPAI models in the legislation constitutes an achievement, especially as their exclusion was pushed for by Europe's "big players", open-source models are presently largely freed from regulatory obligations as long as their do not constitute systemic-risk models. The present classification of systemic risk AI models is anchored at a minimum of 10\string^25 FLOPs which would only capture the most state-of-the-art GPAI models such as Open AI's GPT-4 and Google DeepMind's Gemini \cite{ecqandq}. The exception of open-source providers from transparency requirements alongside the present parameters behind the classification of systemic risk models downplay the significance of the achievement. Open source exceptions that watered down the regulation, largely due to the push by France, Germany, and Italy, provide evidence that Member States actively calculate their downloading costs anchored in maintaining industrial competitiveness and thus try to upload their demands to the EU level in the policy-making process to mitigate future top-down costs. The present exceptions for non-systemic GPAI models leave a window for potential malicious use that can increase risks. Balancing open-source innovation and public safety is not an easy regulatory task but AI safety should not be reduced to strengthen Europe's AI start-up culture. Open-sourced models bear an essential risk as they are widely available to the public. While popular secured AI systems like ChatGPT provide safeguards and decline malicious requests as they violate the firm's usage policy, it is possible to "jailbreak" AI systems and get them to misbehave. Open-source AI systems could be an opening gate for  sophisticated threat actors, who could download the AI system and disable its safety features to adapt it for  malicious purposes \cite{davidevanharrisOpenSourceAIUniquely2024}. It should not be neglected that the main risk stemming out of GPAI models originates from its multi-use, even if the model does not represent the most advanced technology. The open-source release of the LLM "Mixtral 8x7B" \cite{franzenMistralShocksAI2023a} by the French AI firm Mistral in December 2023 illustrates an essential problem linked to open-sourcing AI as the model eclipsed the performance of ChatGPT-3.5  but still falls under the systematic risk threshold.

While the AI Act's open-source GPAI problem constitutes a regulatory patch that should not be overlooked, the legislation sets a prime example on how to mitigate AI risk across several risk domains. However, a prime obstacle in mitigating the global security implications of AI as discussed in \ref{armsracechapter} remains outside the legislation's scope and should be given policy priority: AI in the military and national security setting. The AI Act won't interfere with Member States' sovereignty over national security matters as it won't be subject to systems designed exclusively for military, defense or national security purposes (Article 2, Final Draft). While the legislation's obligations are binding for dual-use technologies, AI systems exclusively used for the purposes listed above remains unregulated which can greatly increase geopolitical risks. While a retrospective integration of exclusive military AI regulation into the Act seems unlikely, its risk-framework could serve as a guiding stick for a military and national-security focused agreement that enables the development of military AI based on shared European values. A European framework for responsible military AI would not only fill a dangerous regulatory gap but signal the bloc's international leadership ambitions in value-based AI governance through mitigating severe risks in both the military and civilian spectrum \cite{fanniWhyEUMust2023}. 

The legislation's risk-based obligatory requirements will have a top-down impact on the Member States that can go beyond the EU's jurisdiction through the Brussels Effect. The shift away from the Commission's definition of AI systems towards an alignment with the position of the OECD internationalizes the bloc's understanding of AI, simplifying future global AI governance built upon a shared conceptual understanding. Maintaining a technological-definitionary alignment with international partners is vital to future-proof the AI Act and to strengthen its potential for a Brussels Effect. It is yet too early to tell if the AI Act has a GDPR-like potential to influence global markets and practices while serving as a blueprint for other jurisdictions looking to regulate AI. With its adoption being completed in March 2024, the legislation is on track to be published in the Official Journal of the EU. The European institutions have displayed an innovative dedication to regulating AI in the first three stages of the policy-cycle; the subsequent pace of implementation and form of enforcement of the mandated obligations will tell if the chosen risk-based approach will be sufficient to mitigate the risks stemming out of AI's disruptive potential.

\section{Conclusion}
"Are you dangerous?" When asking such a simplistic yet complex question to the GPAI model ChatGPT, it does, as expected, not provide an in-depth evaluation of the risks linked to AI. However, it demonstrates why interdisciplinary research on democratized AI, as has been conducted in this work, is necessary:

\begin{quotation}
"As an AI language model, I don't possess intentions or desires, so I'm not inherently dangerous. However, how I'm used or applied can potentially have consequences. It's important for users to employ me responsibly and ethically, as with any tool or technology. Like any powerful tool, the impact depends on how it's utilized by humans. My purpose is to assist and provide information to the best of my abilities while adhering to ethical guidelines" \cite{openaiChatGPT2024}.
\end{quotation}

The response provided by the AI model illustrates that the risk linked to AI allegedly mainly stems from how the technology is employed, emphasizing the importance of responsible and ethical usage. This work's research objective, motivated by the widespread use of AI models such as ChatGPT, was the conceptualization of AI risk from a political perspective and the assessment of the EU's regulatory policy-making capacity concerning AI risk mitigation. Hence, two research questions have been constructed: (RQ1) What kind of risks does AI pose for domestic and international politics? (RQ2) How does the EU AI Act mitigate the risks associated with AI?

The risk taxonomy proposed in Chapter \ref{Chapter 3} identified a total of 12 risks that AI poses for domestic and international politics and categorized them into four domains:(1) Geopolitical Pressures, (2) Malicious Usage, (3) Environmental, Social, and Ethical Risks, and (4) Privacy and Trust Violations. The widespread usage of AI has profound negative implications for political stability, especially if integrated into the defense sector. Disinformation, most commonly achieved by GPAI models that can create deepfakes, is of particular immediate concern given that 2024 stipulates a global election year.

In 2021, the European Commission proposed the Artificial Intelligence Act as a regulatory policy response to the anticipated risk linked to AI. Chapter \ref{Chapter 4} assessed two stages of the EU's policy-cycle, namely policy formulation through the Commission's Proposal and decision-making under the ordinary legislative procedure. The Commission's pioneering Proposal, which differentiated between unaccepted, high-risk, and low-risk AI systems served as a strong foundation for the subsequent policy-making process. However, the spread of GPAI models such as ChatGPT in late 2022 needed an adequate solution by the bloc's lawmakers, as such models' multi-use setting has been difficult to fit into the Proposal. Incorporating GPAI as a separate risk class, as displayed in the Final Draft, which the Parliament and Council adopted in March 2024, is a welcome but incomplete solution to achieve the adoption of the AI Act. Regulation with loopholes is a more welcome outcome than failure and non-agreement would stipulate, especially as the AI Act will have a top-down impact on the EU Member States and holds the potential to go beyond the bloc's direct sphere of influence through a Brussels Effect, potentially positively shaping global AI governance. Yet, existing exceptions for open-source models, excessively high parameters for the classification of GPAI models as a systemic risk, and the exclusion of systems designed exclusively for military purposes from the regulation's obligations leave room for future action. 

Future interdisciplinary research is needed to fill regulatory loopholes in the AI Act and maintain its regulatory capacity to address existing and emerging risks. While the proposed risk taxonomy illustrated the risks AI poses for domestic and international politics, extending it interdisciplinary will be necessary to maintain a sustainable strategy of risk identification. As the policy-making process of the AI Act is about to be completed, future monitoring and compliance enforcement will benefit from research output that breaks down what governance is concerned about, namely the risks that AI poses for our world.

\section*{Acknowledgments}
This paper represents a condensed version of my M.A. thesis, which stems from a two-year research period as a Fulbright and DAAD scholar in the U.S. and the U.K. The research was conducted at the University of North Carolina at Chapel Hill under the supervision of Prof. Dr. Christiane Lemke and was reviewed by Prof. Dr. Gary Marks and Prof. Dr. Liesbet Hooghe.  

\bibliographystyle{unsrtnat}  

\bibliography{references}

\end{document}